 \documentclass[utf8,aps,pre,twocolumn,10]{revtex4-1}

\usepackage{graphicx}
\usepackage{epstopdf}
\usepackage{listings}	
\usepackage{amssymb}
\usepackage{epstopdf}
\usepackage{amsmath}
\usepackage{amsthm}
\usepackage[extension=xxx]{hyperref}
\usepackage{color}
\usepackage{fullpage}
\usepackage[utf8]{inputenc}
\usepackage{algpseudocode}
\usepackage{lipsum}
\usepackage{hyperref}
\definecolor{dkgreen}{rgb}{0,0.6,0}
\definecolor{gray}{rgb}{0.5,0.5,0.5}

\def\be{\begin{equation}}   \def\ee{\end{equation}}
\def\bea{\begin{eqnarray}}   \def\eea{\end{eqnarray}}

\begin{document}
\author{Serge Dmitrieff, François Nédélec}
\title{ConfocalGN : a minimalistic confocal image simulator.}





\begin{abstract}
\textbf{Summary:} We developed a user-friendly software to generate synthetic confocal microscopy images from a ground truth specified as a 3D bitmap with pixels of arbitrary size. The software can analyze a real confocal stack to derivate noise parameters and will use them directly to generate new images with similar noise characteristics. 
Such synthetic images can then be used to assert the quality and robustness of an image analysis pipeline, as well as be used to train machine-learning image analysis procedures.
We illustrate the approach with closed curves corresponding to the microtubule ring present in blood platelets.\\
\textbf{Availability and implementation:}
ConfocalGN is written in {\sc MATLAB} but does not require any toolbox. The source code is distributed under the GPL 3.0 licence on \href{https://github.com/SergeDmi/ConfocalGN}{https://github.com/SergeDmi/ConfocalGN}. 
\end{abstract}

\maketitle

\section*{Introduction}
Confocal microscopy is widely used in the bioscience community, due to its good spatial resolution, and high contrast compared to conventional widefield microscopy. 
However, confocal images suffer from the diffraction limit, detector shot noise, background fluorescence, tagging heterogeneity and fluorophore stochasticity \cite{tsien1995fluorophores,pawley2006fundamental}. 
Moreover, the resolution is anisotropic, and typically several time worse in the direction of the optical axis compared to the other directions. Generally, these effects limit our ability to infer the shape of the objects observed through the microscope. 

Image analysis is the standard approach to extract information from the images. 
It usually involves image processing, such as background subtraction and removal \cite{hwang1995adaptive}, or deconvolution \cite{de2001image}. Additional steps can perform segmentation or directly attempt to reconstruct the objects using a wide variety of techniques, such as model based analysis, feature classification \cite{tomer2010profiling,sommer2011ilastik}, etc.
Image analysis tool tend to include complicated pipelines with adjustable parameters, and for this reason can introduce artifacts with possibly systematic bias \cite{pham2000current}, that may remain unnoticed.  It is thus essential to assess the quality of the analysis procedure, and this can be done by running the analysis pipeline with synthetic images, for which the ground truth is known.  To avoid over-engineering image analysis routines, machine learning is becoming increasingly popular \cite{sommer2011ilastik,van2016deep}. However, machine leaning on 3D images require a large training set,  which could be provided synthetic images with a known ground truth. Both cases require synthetic images presenting  characteristics similar to the real images, and in particular a realistic level of noise.
It is thus extremely useful to be able to easily generate simulated confocal microscope data from a known ground truth. 

\begin{figure*}[t]
\includegraphics[width=\linewidth]{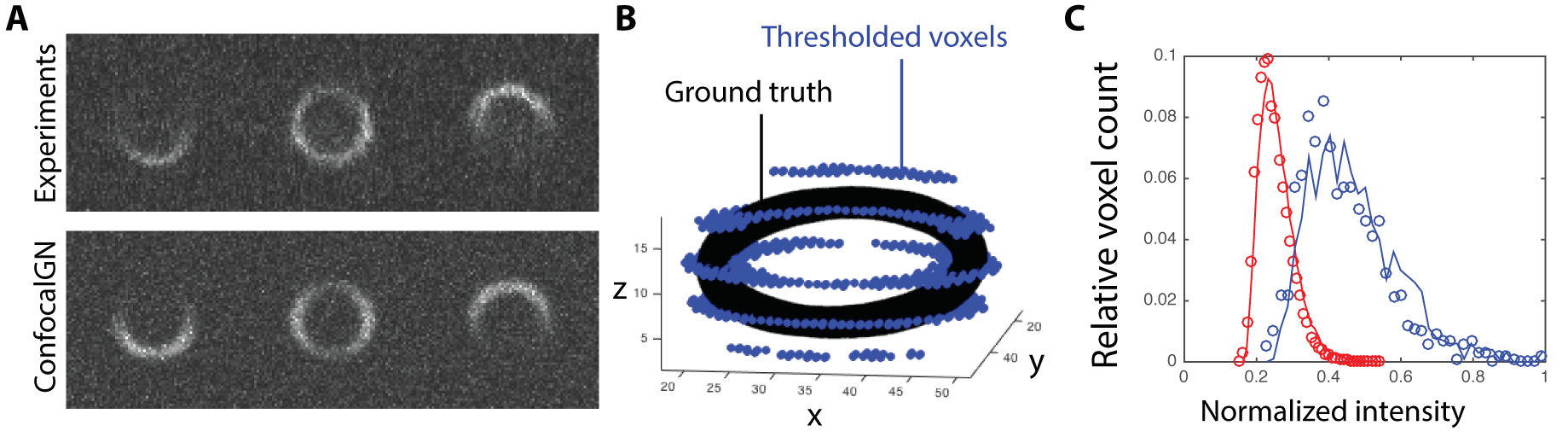}   
\caption{
A, top: images from a real confocal microscopy stack of microtubules in blood platelets \cite{severin2013confocal} (SIR-tubulin data courtesy of A. Mathur). A, bottom: a simulated confocal image of the same. 
B : the ground truth, from which image A bottom was generated is the titled ring (black). The voxels above the Otsu threshold are shown in blue. 
C : histogram of voxel values for the confocal microscopy image (dots) and for a simulated microscopy image (lines).
The pixels have been categorized following the Otsu method, after Gaussian blurring, as noise (red, n=23000 pixels) or signal (blue, n=1000 pixels).
}
\end{figure*}

To simulate a confocal image, one needs to address both optical limitations and noise, {\it i.e.} convolve the ground truth with the point spread function (PSF) of the microscope, and add noise both before and after convolution, according to the nature of the different source of noise. 
Several software (most notably {\em Huygens} or {\em Icy}) can generate synthetic PSF, based on the characteristics of the microscope components, i.e. the optical path. Possible limitations include the non-uniformity of the refraction index in real samples and the aberrations that could stem from lens misalignment and imperfections. It is also possible to directly measure the {\em experimental} PSF from microscopy images. 

Starting from a known PSF, a couple of software can simulate confocal imaging: {\em Huygens software professional} and the open-source {\em microlith}. While the later does not include a noise simulator, the former allows the user to add a background as well as a Gaussian or Poisson noise of specified level; however, knowing the correct noise intensity and distribution is not straightforward. Each source of noise (background noise, autofluorescence, shot noise) could be simulated \cite{sheppard2006signal}, but the operation involves the convolution of the different sources of noise and distortions with the PSF, and can only be computer accurately in simplified experimental conditions \cite{herberich2012signal}. 

Practically, any confocal simulator will be limited to a certain level of details, and by the lack of information on the sample structure and properties - which is precisely the unknown. While such detailed simulators are extremely useful, we followed here a simpler approach, which is sufficient to test an image analysis pipeline.



\section*{Results}
We developed ConfocalGN, a minimal solution to simulate confocal datasets from a ground truth, a measure of the PSF, and the noise and signal distribution. 

\begin{itemize}  
\item The ground truth is provided as a high resolution image, either as a {\sc MATLAB} matrix, or as a TIFF file. The value of the pixels correspond to the level of fluorescence, with $0$ indicating no fluorescence. Pixel values are rescaled to match the desired mean signal value. The pixel size can be arbitrary and is specified as a parameter.
\item The PSF is approximated by a Gaussian, and is thus specified as a three components vector containing the standard deviation in each direction (in units of the ground truth image pixels). Confocal PSF are usually well approximated by Gaussians \cite{zhang2007gaussian}. We do not provide a mean to compute the parameters of the PSF, but several third party software offer this possibility.
\item The level of noise is specified as a vector of the three first moments of the noise distribution, which can readily be measured from a sample experimental image. The synthetic noise is generated using is the gamma distribution, which is a good approximation of reality \cite{herberich2012signal}. Note that a Gaussian approximation is used if the user-provided noise has no skew ({\it i.e.} if the third moment is zero).
 \item The signal intensity is entered as the mean value of the pixels, which can be observed experimentally.
 \end{itemize}

The ground truth images will usually be generated numerically using various software. 
Importantly, ConfocalGN offers the possibility to calculate the noise and signal intensity directly from a sample image. 
Thus by specifying one input image, the user can obtain a realistic noise without analyzing the noise and distortion sources. 
 
ConfocalGN offers a minimal segmentation function doing gaussian blurring and Otsu thresholding to discriminate noise vs signal pixels, but the user can provide his own segmentation function instead. In Fig. 1A, a real (top) and simulated image (bottom) are presented. Fig.1C, shows the histogram of voxel values after the image has been segmented in two groups by Gaussian  blurring plus Otsu thresholding : noise (red) and signal (blue). 
 
In conclusion, ConfocalGN is simple to use and useful to test image analysis pipelines. 
It is not meant to be an exact simulation, and adopts commonly accepted approximations of the PSF and simplified treatments of fluorophore emission, background and sensor noise. 
The user is thus only required to specify: (1) a ground truth image, (2) three characteristics of the PSF, and (3a) a sample image or (3b) the moments of the noise and signal distributions. 
Starting from measured PSF and noise characteritics it can robustly and efficiently produce simulated confocal image stacks with realistic characteristics.

\section*{Acknowledgements}
The authors thank Aastha Mathur for the confocal microscopy data and scientific comments, Joran Deschamps for the constructive comments on the manuscript and the EMBL image analysis community for the scientific suggestions.

\bibliographystyle{unsrt}
\bibliography{bibtex_refs}

\begin{thebibliography}{10}

\bibitem{tsien1995fluorophores}
Roger~Y Tsien and Alan Waggoner.
\newblock Fluorophores for confocal microscopy.
\newblock In {\em Handbook of biological confocal microscopy}, pages 267--279.
  Springer, 1995.

\bibitem{pawley2006fundamental}
James~B Pawley.
\newblock Fundamental limits in confocal microscopy.
\newblock In {\em Handbook of biological confocal microscopy}, pages 20--42.
  Springer, 2006.

\bibitem{hwang1995adaptive}
Humor Hwang and Richard~A Haddad.
\newblock Adaptive median filters: new algorithms and results.
\newblock {\em IEEE Transactions on image processing}, 4(4):499--502, 1995.

\bibitem{de2001image}
Jacques~Boutet De~Monvel, Sophie Le~Calvez, and Mats Ulfendahl.
\newblock Image restoration for confocal microscopy: improving the limits of
  deconvolution, with application to the visualization of the mammalian hearing
  organ.
\newblock {\em Biophysical Journal}, 80(5):2455--2470, 2001.

\bibitem{tomer2010profiling}
Raju Tomer, Alexandru~S Denes, Kristin Tessmar-Raible, and Detlev Arendt.
\newblock Profiling by image registration reveals common origin of annelid
  mushroom bodies and vertebrate pallium.
\newblock {\em Cell}, 142(5):800--809, 2010.

\bibitem{sommer2011ilastik}
Christoph Sommer, Christoph Straehle, Ullrich K{\"o}the, and Fred~A Hamprecht.
\newblock Ilastik: Interactive learning and segmentation toolkit.
\newblock In {\em 2011 IEEE international symposium on biomedical imaging: From
  nano to macro}, pages 230--233. IEEE, 2011.

\bibitem{pham2000current}
Dzung~L Pham, Chenyang Xu, and Jerry~L Prince.
\newblock Current methods in medical image segmentation 1.
\newblock {\em Annual review of biomedical engineering}, 2(1):315--337, 2000.

\bibitem{van2016deep}
David~A Van~Valen, Takamasa Kudo, Keara~M Lane, Derek~N Macklin, Nicolas~T
  Quach, Mialy~M DeFelice, Inbal Maayan, Yu~Tanouchi, Euan~A Ashley, and
  Markus~W Covert.
\newblock Deep learning automates the quantitative analysis of individual cells
  in live-cell imaging experiments.
\newblock {\em PLOS Comput Biol}, 12(11):e1005177, 2016.

\bibitem{severin2013confocal}
Sonia Severin, Fr{\'e}d{\'e}rique Gaits-Iacovoni, Sophie Allart, M-P Gratacap,
  and Bernard Payrastre.
\newblock A confocal-based morphometric analysis shows a functional crosstalk
  between the actin filament system and microtubules in thrombin-stimulated
  platelets.
\newblock {\em Journal of Thrombosis and Haemostasis}, 11(1):183--186, 2013.

\bibitem{sheppard2006signal}
Colin~JR Sheppard, Xiaosong Gan, Min Gu, and Maitreyee Roy.
\newblock Signal-to-noise ratio in confocal microscopes.
\newblock In {\em Handbook of biological confocal microscopy}, pages 442--452.
  Springer, 2006.

\bibitem{herberich2012signal}
Gerlind Herberich, Reinhard Windoffer, Rudolf~E Leube, and Til Aach.
\newblock Signal and noise modeling in confocal laser scanning fluorescence
  microscopy.
\newblock In {\em International Conference on Medical Image Computing and
  Computer-Assisted Intervention}, pages 381--388. Springer, 2012.

\bibitem{zhang2007gaussian}
Bo~Zhang, Josiane Zerubia, and Jean-Christophe Olivo-Marin.
\newblock Gaussian approximations of fluorescence microscope point-spread
  function models.
\newblock {\em Applied Optics}, 46(10):1819--1829, 2007.

\end{thebibliography}

\end{document}